\documentclass{article}
\usepackage[square,numbers]{natbib}

\usepackage[preprint]{neurips_data_2024}

\usepackage[utf8]{inputenc} 
\usepackage[T1]{fontenc}    
\usepackage{hyperref}       
\usepackage{url}            
\usepackage{booktabs}       
\usepackage{amsfonts}       
\usepackage{nicefrac}       
\usepackage{microtype}      

\usepackage[table]{xcolor}

\usepackage{multirow}
\usepackage{makecell}
\usepackage{adjustbox}
\usepackage{xspace}

\usepackage{array}    
\usepackage{graphicx}
\usepackage{subcaption} 
\usepackage{tabularx}   

\newcolumntype{P}[1]{>{\raggedleft\arraybackslash}p{#1}}

\newcommand{\paratitle}[1]{\smallskip\noindent\textbf{#1}}
\newcommand{\ie}{\emph{i.e.,}\xspace}
\newcommand{\eg}{\emph{e.g.,}\xspace}

\newcommand{\dataset}{TVR-Ranking\xspace}

\usepackage{amsmath}

\title{TVR-Ranking: A Dataset for Ranked Video Moment Retrieval with  Imprecise  Queries}

\author{%
 Renjie Liang\textsuperscript{1},\quad Li Li\textsuperscript{1},\quad Chongzhi Zhang\textsuperscript{1},\quad Jing Wang\textsuperscript{1},\quad  Xizhou Zhu\textsuperscript{2},\quad   Aixin Sun\textsuperscript{1}\\
\textsuperscript{1}Nanyang Technological University\\
\textsuperscript{2}SenseTime
}

\begin{document}

\maketitle

\begin{abstract}
   In this paper, we propose the task of \textit{Ranked Video Moment Retrieval} (RVMR) to locate a ranked list of matching moments from a collection of videos, through queries in natural language. Although a few related tasks have been proposed and studied by CV, NLP, and IR communities, RVMR is the task that best reflects the practical setting of moment search. To facilitate research in RVMR, we develop the TVR-Ranking dataset, based on the raw videos and existing moment annotations provided in the TVR dataset. Our key contribution is the manual annotation of relevance levels for 94,442 query-moment pairs. We then develop the $NDCG@K, IoU\geq \mu$ evaluation metric for this new task and conduct experiments to evaluate three baseline models. Our experiments show that the new RVMR task brings new challenges to existing models and we believe this new dataset contributes to the research on multi-modality search. The dataset is available at \url{https://github.com/Ranking-VMR/TVR-Ranking}
\end{abstract}

\section{Introduction}
\label{sec:intro}
Given a query expressed in natural language, to retrieve or to locate a temporal moment from video(s) that semantically matches the query has many applications. A temporal moment refers to a segment within a source video with identified start and end timestamps. Examples of such applications include searching for a specific scene in security surveillance videos~\cite{yuan2023surveillance}, locating a medical procedure within an educational tutorial~\cite{gupta2023dataset}, or identifying desired scenes for video editing purposes, among others. 

\begin{figure}
    \centering
    \includegraphics[width=0.95\columnwidth]{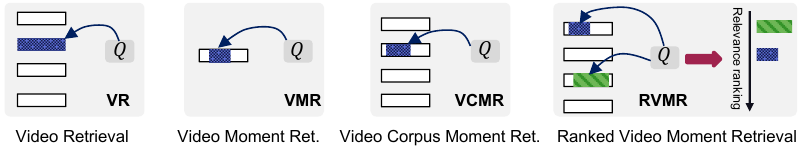}
    \caption{RVMR and its related tasks. A rectangle represents a video; the matching moment to text query $Q$ is shaded. In VR VMR and VCMR, exactly one video/moment is to be retrieved.}
    \label{fig:taskComparison}
\end{figure}

A few tasks with different names have been studied for addressing similar objectives, including video retrieval (VR), video moment retrieval (VMR), natural language video localization (NLVL), temporal sentence grounding in video (TSGV), and video corpus moment retrieval (VCMR)~\cite{Liu2023VLMSurvey,Zhang2023TSGVSurvey}. Among them, VR involves retrieving a video from a collection based on visual content, akin to video search on platforms like YouTube, but with the search criteria grounded in the visual content of videos. NLVL and TSGV, more commonly used in CV and NLP communities, refer to the same task as video moment retrieval (VMR) in the IR community. VMR aims to locate a moment within a given video that semantically matches the text query. The VCMR task is a direct extension of VMR, focusing on retrieving a moment from a collection of videos~\cite{victor2019temporal}. As depicted in Figure~\ref{fig:taskComparison},  existing tasks VR VMR and VCMR, all aim to find one answer, being either a video or a moment,  for a given query. 

The reason for expecting  one exact answer to a query in existing datasets lies primarily in the annotation of benchmark datasets. During  annotation, annotators watch a video, then provide textual descriptions of meaningful video moments in this video.  Subsequently, each description serves as the query to retrieve the corresponding moment from this source video. Given that a query typically describes a specific moment precisely, a model trained on these datasets can assume the existence of the moment to be searched for, and all queries are from users who possess a good understanding of the source video. 

In a practical setting, there exist multiple moments that can be described similarly, even in a single video.  For example, one video may contain multiple moments for ``Phoebe enters room and sits on sofa'', or a very similar moment ``Alice enters room and sits on sofa''. If we assume a user has limited knowledge about the source video, then he/she may formulate a query like ``a woman enters room and sits on sofa''. In this case, all the moments that correspond to either Phoebe or Alice entering a room and sitting on sofa are perfect matches. Further, there could be relevant but non-perfect matches like ``a man enters room''. 

In this paper, we define a new task named \textit{Ranked Video Moment Retrieval} (RVMR) to better reflect the practical setting. \textbf{RVMR} is to retrieve a ranked list of moments matching an \textit{imprecise} query from a collection of videos. We do not assume that users have fully watched all videos to be searched. Hence, users may or may not provide a precise description of a specific moment. Accordingly, multiple moments from the same or different videos can be retrieved and ranked by their degrees of relevance to the query. Compared to exisisting tasks, RVMR exhibits two distinct characteristics: (i) not all queries provide precise descriptions of moments, and (ii) retrieved moments are ranked by their relevance to the query. While models designed for VCMR can potentially be repurposed for RVMR, they may lack the necessary moment ranking capability.

To date, no datasets cater to this novel task setting. Thus, we have curated the \textbf{\dataset} dataset to facilitate the RVMR task. As its name suggests, the dataset has its root in the TVR \footnote{The TVR dataset is under the MIT License.} dataset~\cite{lei2020tvr}. Specifically, we reuse the source videos in TVR, \ie video clips from six TV series, and the original moment annotations \ie the begin/end timestamps of meaningful moments. We have made two main efforts in the development of the \dataset. The first is \textit{deriving imprecise queries} from the original moment descriptions. In the TVR dataset, moment descriptions are very detailed and often contain TV character names. We substitute  character names with pronouns using carefully crafted prompts to ChatGPT, with follow up quality control. This process entails replacing a total of 160,701 words across 72,842 moment descriptions. Our second effort involves \textit{annotating the relevance scores of moments} to these imprecise queries. This task was undertaken by 23 annotators over 1,200 working hours. We have annotated relevant moments for 3,281 queries. Among them, 52\% of queries were each annotated with 20 candidate moments, featuring five relevance levels ranging from irrelevant (0) to a perfect match (4). The remaining 48\% of queries were each annotated with 40 candidate moments. The annotation of the relevance score for a candidate moment to a query (\ie a query-moment pair) was conducted by either two annotators (in case of consensus can be reached by the two) or four annotators. A total of 94,442 relevance scores were annotated with consensus. The annotated queries are divided into 500 validation and 2,781 test queries.

Our contributions in this resource paper are summarized as follows. First, we \textbf{define the RVMR task} to reflect the practical scenario of retrieving moments from video collections using imprecise queries. Second, we have \textbf{annotated the \dataset} which provides 3,281 queries and their relevance annotations, each with 20 or 40 candidate moments. Third, we propose an \textbf{evaluation metric} for the new task named $NDCG@K, IoU \geq \mu$. This metric builds upon $NDCG$, designed for ranking tasks, by incorporating the $IoU$ metric to handle partial matches between the retrieved and the ground truth moments. Lastly, we \textbf{adapt three baseline models} (initially designed for VCMR) to RVMR, and evaluate their performance on the \dataset. Our results suggest that models effective on VCMR may not perform well on RVMR. 

On top of these contributions, we believe that the \dataset provides new opportunities to explore video/moment indexing mechanisms for both efficient and effective video moment retrievals.

\section{Related work}
\label{ssec:related}

\begin{table*}[t]
    \caption{Existing VMR datasets and our annotated TVR-Ranking dataset. `M.Dur' and `V.Dur' mean the total duration of moments and videos, respectively, in hours.  In TVR-Ranking, we select the top $N=40$ moments by query-caption similarity for a query, to comprise the pseudo training set. In the annotated validation and test sets, the average number of relevant moments per query is 27. For all other datasets, the matching moment per query is 1.0 except QVHighlight which is 1.78.}
    \label{tbl:existing_datasets}
   \footnotesize
    \begin{tabular}{l|rrr|rr|rr}
        \toprule
        Dataset & {\#Query} & {\#Moment} & {\#Video}  &  {Vocab.}   & {\#Verb} & M.Dur  & V.Dur \\
        \midrule
        Charads-STA~\cite{gao2017tall}  & 16,128  & 11,770 & 6,672  & 1,303 & 469 & 26.33 & 56.69 \\
        ActivityNet Captions~\cite{krishna2017dense}   & 54,559  & 54,559 & 14,926  & 13,645 & 4,510 & 560.37 & 487.60 \\ 
        TACoS ~\cite{regneri2013grounding}  &  18,227  & 7,069 & 127 & 2,287 & 994 & 2220.82 & 10.11 \\
        MAD (v2-unnamed)~\cite{soldan2022mad}     & 3,328,745 & 328,742 & 488  & 56,066 & 13,993 & 263.46 & 1,207.3 \\
        Ego4D(NLQ)~\cite{Grauman_2022_CVPR}  & 18,399  & 18,374 & 1,685 & 3,337 & 823 & 56.85 & 231.82    \\
        UCA~\cite{yuan2023surveillance}    &  19,211 & 18,299 & 1,544  & 4,087 & 1,580 & 84.39 & 96.71 \\
        MedVidQA~\cite{DBLP:journals/corr/abs-2201-12888} & 3,010& 2,990 & 899  & 2,291 & 670 & 51.78 & 95.72 \\
        QVHighlight~\cite{lei2021detecting}    &  10,310 & 18,367 & 10,148 &    7,750 & 1,824 & 125.50 & 422.83 \\
        TVR~\cite{lei2020tvr}  & 98,070  & 97,442 & 19,614  & 18,856 & 6,104 & 240.70 & 414.86  \\
        \midrule 
        TVR-Ranking (Training)  & 69,317   & 94,259 & 19,614  & 10,865 & 3,846 & 228.91 & 414.86 \\ 
        TVR-Ranking (Validation)  & 500  & 12,191 & 9,272  & 994 & 330 & 29.48 & 197.27   \\
        TVR-Ranking (Test Set)  & 2,781 & 42,472  & 18,146   & 2,517 & 837 & 101.60 & 384.23 \\
        \bottomrule
    \end{tabular}    
\end{table*}

Current VMR and VCMR datasets fail to simulate real-world moment search scenarios due to two key unrealistic assumptions: users have a deep understanding of the source video and there is only one "perfect match" moment for each query.

The first assumption stems from traditional annotation processes where annotators are required to watch the entire video and then describe meaningful moments therein. Most datasets listed in Table~\ref{tbl:existing_datasets} are annotated in this way, including  DiDeMo~\cite{Hendricks_2017_ICCV},  TACoS~\cite{regneri2013grounding},  TVR~\cite{lei2020tvr}, ActivityNet Caption~\cite{krishna2017dense}, Ego4D(NLQ)~\cite{Grauman_2022_CVPR}, and UCA~\cite{yuan2023surveillance}. Besides, the Charades-STA dataset~\cite{gao2017tall} extends the Charades dataset~\cite{sigurdsson2016hollywood} by segmenting video descriptions into sentences and linking them to specific video timestamps via keywords. In contrast, queries in our dataset may or may not provide precise descriptions of moments, thus embracing users with different levels of understanding of the corpus.

Typically, standard VMR datasets generally link a query to a single relevant moment.  For instance, the health-related queries in the MedVidQA dataset~\cite{DBLP:journals/corr/abs-2201-12888}, sourced from WikiHow’s ‘Health’ category, are well-represented of real-world scenarios. Nonetheless, this dataset confines each query to just the most relevant moment. In contrast, real-life situations frequently encompass multiple moments that can be similarly described. The QVHighlight dataset~\cite{lei2021detecting} was pioneering in allowing queries to match multiple moments within a single video. However, it still restricts searches to single videos and focuses only on perfect matches. Our dataset aims to retrieve a ranked list of moments from a video corpus based on imprecise queries, functioning more like a search engine and accommodating both closely and loosely relevant matches. This paradigm not only broadens the utility of the results but also aligns more closely with practical search needs.

\section{The TVR-Ranking Dataset Annotation}
\label{sec:annotation}

In the ideal setting, a dataset shall well reflect the context of real-world applications, \eg the data source and the information needs from users~\cite{Yu2023dataset}. In the RVMR task setting, we assume there exists a collection of videos, and users search for relevant moments through textual descriptions as queries. However, such kinds of queries can only be collected from logs of video search services, which are not publicly accessible. Without access to such resources, we choose to derive user queries from existing data annotations, \ie the datasets listed in Table~\ref{tbl:existing_datasets}.

Our immediate task is to choose which existing dataset to use as the raw data for annotation. 
To this end, we compare the existing datasets based on the following perspectives: accessibility to the raw videos, number of videos, variants of different activities/scenes, and number of moments annotated. Because existing annotations are mostly descriptions of scenes and/or actions, the number of verbs  has been widely used to measure the number of activities covered in a dataset~\cite{sigurdsson2017actions}. Based on these considerations, we adopt the TVR dataset as the raw source for our annotation; accordingly, our dataset is named \textbf{TVR-Ranking}. 

The TVR dataset contains video clips from six different TV series, along with their corresponding subtitles and audio tracks. 
In the construction of the original TVR dataset, annotators were tasked with identifying the boundaries of events \ie moments, within these clips, and describing their content. 
Workers also provided whether the description was purely based on the visual content, the subtitle, or both the video and subtitle. The TVR dataset comprises 72,842 video-only, 8,920 subtitle-only, and 16,308 video-subtitle moment descriptions.\footnote{The numbers reported here are from the dataset version used in our annotation, with negligible differences from those reported in the original paper~\cite{lei2020tvr}. }  In our TRV-Ranking construction, we aim to concentrate on the visual aspects; therefore, we only consider the annotations that are purely based on the visual content of the videos.

 \subsection{Imprecise Queries}
 \label{ssec:query_collection}

In TVR dataset, many moment descriptions contain character names and even their dressing details, making them precise descriptions of the moments. To derive imprecise queries, we substitute these words with more general words. In particular, we replace all character names with pronouns.  This replacement is essential for our annotation because our annotators (also users) may not have knowledge about these characters. Table~\ref{tab:queryReplacement} lists three example descriptions before and after substitution. 

\begin{table}
    \centering
    \small
    \caption{Three example moment descriptions before and after word substitution. }
    \label{tab:queryReplacement}
    \begin{tabular}{c|p{2.6in}|p{2in}}
    \toprule
     No. & Original query before word substitution & Query after word substitution \\ \midrule
    1.& \textit{Eric and Dr. Gregory} were having a conversation.                 
        & Two people were having a conversation.  \\  
     2.&   \textit{Rachel Green and Ross} were having a conversation.  
       &  Two people were having a conversation.  \\ 
      3.& \textit{Javier and the young man wearing checkered polo} was having a conversation. 
         & Two people were having a conversation. \\
    \bottomrule 
    \end{tabular}
\end{table}

The character name substitution is through carefully designed prompts to ChatGPT (see Appendix~\ref{appendix:replacement_implementation}), with quality checks. The output of ChatGPT is a quality substitution if it successfully passes two validation checks. The first check is for semantic consistency, to ensure no significant change in terms of semantic meaning after substitution. The SimCSE~\cite{gao2021simcse} similarity of moment descriptions before and after the substitution is expected to be above a threshold (0.4 in our implementation). The second check is to ensure no person names appear in the substituted version. We detect person names in the substituted moment description using Flair~\cite{akbik2019flair}. If a substitution fails to pass both checks, the moment description undergoes human review and is fed to ChatGPT again for another substitution with a different temperature parameter setting, till it passes the two checks.

The above procedure replaces 160,701 words across 72,842 moment descriptions, averaging 2.21 words per description.
Table~\ref{table:replaced_word} presents the top 12 most frequently replaced words, which shows a diversity of personal pronouns in the descriptions.\footnote{ 
The implementation details and the prompt are detailed in Appendix~\ref{appendix:replacement_implementation}}

\begin{table}
    \centering
    \small
    \caption{The top 12 most frequent replacement words. M, F, and N denote male, female, and gender-neutral, respectively.}
    \label{table:replaced_word}
    \begin{tabular}{l|rc||l|rc}
    \toprule 
        Word    & Freq. & Gender    & Word      & Freq.    & Gender            \\  \midrule
        man     &  37,533    & M         & some      & 1,747         & N    \\  
        woman   &  34,708    & F         & doctor    & 1,263         & N    \\ 
        person  &  28,109    & N         & other     & 1,149         & N    \\ 
        two     &  6,426     & N         & they      & 1,064         & N    \\ 
        people  &  6,035     & N         & her       &  572          & F  \\ 
        someone & 1,938      & N         & guy       &  533          & M              \\ 
        \bottomrule
    \end{tabular}
\end{table}

To distinguish the moment descriptions before and after the substitution, we call the substituted version \textbf{moment caption}. Observe in Table~\ref{tab:queryReplacement}, the three moment descriptions become the same moment caption after substitution. There are two implications. One is that the moment caption is imprecise, achieving our design goal. The other is that there are likely multiple moments matching the same imprecise query. Accordingly, we use the imprecise moment captions as queries in our TVR-Ranking dataset. We also specifically deal with the identical queries in our annotation, to avoid unnecessary repetition.

\subsection{Relevance Annotation and Quality Control}
\label{ssec:queryCollection}

From the 72,842 moment captions, we randomly select 500 and 2781 moment captions as queries in validation and test sets respectively, for manual annotation. The remaining moment captions will be used to construct a pseudo training set, to be detailed in Section~\ref{ssec:training_generation}. As a search task, all queries share the same large pool of source videos.

Next, we manually annotate ground truth moments, along with their degree of relevance, for both the 500 validation and 2781 test queries. Manually annotating all matching moments from such a large video corpus for a given query is infeasible. Therefore, we utilize the moment annotations available in the original TVR dataset during the annotation process. These original annotations serve two purposes. 
First, we fully rely on the temporal boundaries of all moments in the original TVR annotations. This approach allows us to view the video corpus as a vast collection of moments each accompanied by a moment caption, during the data annotation process. Second, the moment caption provides a reasonably good description of a moment. Consequently, the semantic similarity between a query and a moment caption acts as a proxy for an initial estimation of their relevance.

Let $m.c$ and $m.v$ represent the caption and the visual content of moment $m$, respectively. To annotate the ground truth moments for a query $q$, we initially retrieve the top-$K$ moment candidates based on their similarity to $q$ by using SimCSE~\cite{gao2021simcse}, denoted by using $sim(q, m.c)$. In our annotation process, we set $K=20$ for the first batch.
This batch of 20 query-moment pairs is then presented to two annotators.\footnote{The query $q$ is the same in the batch of 20 query-moment pairs $\langle q, m.v\rangle$. However, during annotation, an annotator is presented with one query-moment pair each time through the annotation interface.}  Each annotator independently labels the degree of relevance of every query-moment pair $\langle q, m.v\rangle$ purely based on the moment's visual content, assigning a score from 0 for irrelevant, to 4 for a perfect match.  

If the difference between the two relevance scores assigned by two annotators is either 1 or 0, then we considered the two annotators to have reached a consensus. The average relevance score was then rounded up to the nearest whole number as the final score for this query-moment pair. If the two annotators fail to reach a consensus, the same query-moment pair is assigned to another two annotators. Then we have a total of 4 scores. Among the 4 scores, we remove one highest score and one lowest relevance score. If the difference between the remaining two scores is either 1 or 0, we consider a consensus is reached;  the average of the remaining two scores is rounded up to the nearest whole number as the final score. Otherwise, the pair is discarded.

After completing the annotation of all 20 moments for a query in the first batch, the lead annotator (the first author) checks the relevance score distribution of these 20 candidate movements. Recall that the 20 candidate movements are ranked by $sim(q,m.c)$. Movements ranked in the top few positions are likely to be more relevant than those ranked lower. However, if the last 5 candidates among the 20 remain very relevant, then it is a strong indication that the annotation so far has not fully covered all matching moments. The next batch of 20 candidate moments will be retrieved by $sim(q,m.c)$ for annotation. We observe that we can cover all relevant movements for nearly every query after annotating the second batch, totaling 40 candidate moments. Hence, at most, two batches or 40 candidate moments are annotated for a query.

At the completion of the annotation process, we obtained a total of 9,272 valid annotations for the 500 validation queries and 18,146 annotations for the 2,781 test queries. This resulted in a total annotation cost of approximately $13,000$ USD for around 1,200 working hours contributed by 23 annotators, excluding the lead annotator's effort. All annotators underwent a tutorial and qualifying exam before participating in the annotation task, as detailed in Appendix~\ref{appendix:annotator}.

 \subsection{Pseudo Training Set Generation}
 \label{ssec:training_generation}

Due to the high annotation cost, we do not manually annotate training data. Instead, we rely on the query-caption similarity, i.e., $sim(q, m.c)$, as a proxy to generate pseudo annotations as the training set. Specifically, given a query, we collect the top-$N$ moments based on $sim(q, m.c)$ as the training set. In our dataset, we include pseudo training sets with $N=1$, $N=20$, and $N=40$. Datasets with other values of $N$ can be easily generated as well. 

As shown in Table~\ref{tab:queryReplacement}, after the substitution, two moment descriptions may become identical. To ensure all queries in the validation and tests do not appear in training, we remove from the pseudo training set the queries that appear either in validation or test, a total of 244 queries. As a result, the pseudo training set contains a total of 69,317 queries.

\subsection{TVR-Ranking: Statistics}
\label{ssec:dataStatis}

\begin{table}
\centering
\caption{Statistics of the TVR-Ranking Dataset. The average ratio of the moment duration to the entire video duration}
    \label{tab:DatasetStatistic}
    \footnotesize
    \begin{tabular}{l |r r r}
    \toprule
         \multicolumn{2}{r}{ \textbf{Pseudo training set,  $N$=40 }} & \textbf{Validation Set} & \textbf{Test Set} \\
    \midrule
    Min Query Length & 4 & 7 & 6  \\
    Avg. Query Length & 13.98 & 14.11 & 13.97  \\ 
    Max Query Length & 122 & 35 & 108  \\ \midrule
    Min Moment Duration (s) & 0.26 & 0.27 & 0.26 \\
     Avg. Moment Duration (s) & 8.74 & 8.71 & 8.61 \\ 
     Max Moment Duration (s) & 239.38 & 121.86 & 138.02  \\ \midrule
    Min Video Duration (s) & 2.02 & 2.02 & 2.02 \\
    Avg. Video Duration (s) & 76.14 & 76.59 & 76.23 \\   
    Max Video Duration (s) & 272.02 & 272.02 & 272.02  \\\midrule  
    Avg. Moment-Video Duration Ratio\textsuperscript{1} & 0.12 & 0.11 & 0.11 \\
    
    Avg. Relevant Moments per Query & N.A & 27.1 & 27.0  \\
    \bottomrule
    \end{tabular}
\end{table}

Table~\ref{tab:DatasetStatistic} provides an overview of the annotated TVR dataset, with query length in number of words, moment duration in seconds, and the source video duration in seconds. The average ratio of moment length to its source video length is about one-tenth. In the table, we also list the average number of relevant moments (with a relevance score of 1 to 4) annotated per query is around 27. Recall that, moments are annotated in batches to a query, with each batch containing 20 candidate moments. Specifically, in the validation set, 264 queries (52.80\%) were annotated with 20 moments (\ie one batch) and 236 queries (47.20\%) with 40 moments (\ie two batches). The test set follows a similar distribution, with 1,473 queries (52.97\%) annotated with 20 moments and 1,308 queries (47.03\%) with 40 moments. The justification of annotating at most 40 moments for a query is detailed in Appendix~\ref{appendix:annotationAnalysis}. 

Following the consensus verification process, 14,382 (97.79\%) annotations in the validation set reached consensus with either two or four annotators, while 325 (2.21\%) were found to be in disagreement and subsequently discarded. Each annotation here is a query-moment pair. Again, the test set shows a similar distribution; 80,060 (97.90\%) annotations achieved consensus and 1,716 (2.10\%) annotations were discarded. With the annotations in consensus, on average each query comes with 27.1 relevant moments in the validation set, and 27.0 in the test set, by counting the moments with relevance scores from 1 to 4.

 \section{Evaluation Metric for RVMR}
 \label{sec:vmr_ranking_metric}

For RVMR, we aim to retrieve a ranked list of relevant moments for a given text query from a video collection. The quality of retrieval can be evaluated from at least two aspects: (i) the quality of moment localization, \ie to what extent the model correctly identifies the temporal boundaries of a moment, and (ii) the quality of ranking, \ie to what extent the model correctly ranks the retrieved moments from most to least relevance to the query. Note that, even if a moment is correctly located with perfect start/end timestamps, the moment may not be the more relevant to the query.

\begin{figure}
    \centering
    \includegraphics[width=0.9\columnwidth]{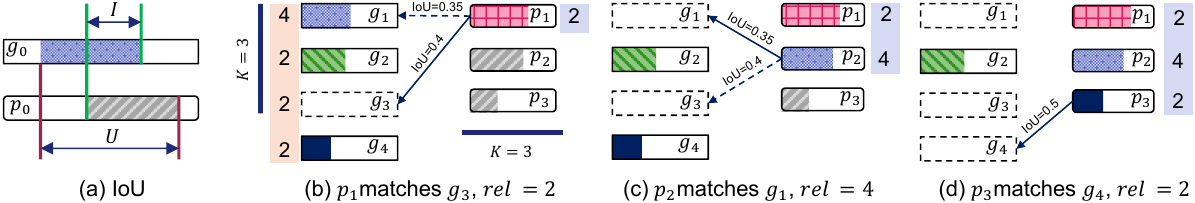}
    \caption{Illustration (a) IoU, and (b)--(d) for $NDCG@3, \mu=0.3$. (b) $p_1$ matches $g_3$ with $rel=2$ for the larger $IoU$, above the 0.3 threshold. (c) $p_2$ matches $g_1$ as $g_3$ is no longer available. (d) $p_3$ matches $g_4$, with $rel=2$.} 
    \label{fig:ndcgIoUAtk}
\end{figure}

\paratitle{$IoU$ for Moment Localization}. Intersection over Union ($IoU$), denoted by $\mu$, is a common metric widely used in moment retrieval tasks. Given a moment prediction with start and end timestamps, evaluated against the ground truth start and end timestamps, IoU measures the intersection along the timeline against the union along the timeline, illustrated in Figure~\ref{fig:ndcgIoUAtk}(a), where $g_0$ and $p_0$ denote the ground truth and predicted moments respectively. If there is no overlap between the two moments, then $\mu=0$. $IoU$ is commonly used as pre-selection criteria for qualifying moments before other measures are computed. For example, a model can be measured by the ability to locate moments with $\mu\geq 0.3$.

\paratitle{\textit{NDCG} for Ranked Retrieval}. Normalized Discounted Cumulative Gain (\textit{NDCG}) is delicately designed for evaluating ranking results with different relevance levels~\cite{manning2008introduction}. Specifically, the Discounted Cumulative Gain (\textit{DCG}) of the top $K$ ranked results is defined in Equation~\ref{eqn:dcg}, where $i$ is the ranking position with 1 being the top ranked position, $rel_i$ is the level of relevance. For example, the left part of Figure~\ref{fig:ndcgIoUAtk}(b) shows four ground truth moments with $g_1$ at rank 1 position and $g_4$ at rank 4 position. To their left are the relevance levels with $rel_1=4$ for $g_1$ and $rel_2=2$ for $g_2$.
\begin{equation}
    DCG@K=\sum_{k=1}^K\frac{rel_i}{log_2(i+1)}
    \label{eqn:dcg}
\end{equation}
$NDCG@K$ is then defined as the \textit{normalized} $DCG$ against the $DCG@K$ value of a perfect ranking \eg all items with a relevance level of 4 are ranked before all items with a relevance level of 3, and so on, till the $K$ cut.

\paratitle{The Proposed Metric $NDCG@K,IoU\geq \mu$ for RVMR.} We process the matching of predicted moments following their ranking returned by a model. If a predicted moment fails to find a matching ground truth with an $IoU\geq\mu$, it is  assigned a relevance score of 0. When multiple ground truths meet the $IoU\geq \mu$ criterion, we select the one with the highest $IoU$ and remove it from the ground truth moment listing, to prevent duplicate matches. The $NDCG@K$ is computed by the relevance scores of the predicted moments at cut $K$, and the perfect ranking of the top $K$ ground truth moments, regardless these $K$ moments are matched by any predicted moment or not.

Figure~\ref{fig:ndcgIoUAtk} shows the computation of $NDCG@3, IoU \geq 0.3$ as an example. Starting with top predicted moment $p_1$, where $IoU(p_1, g_1)=0.35$ and $IoU(p_1, g_3)=0.4$, both exceeding the threshold $\mu=0.3$, as shown in Figure~\ref{fig:ndcgIoUAtk}(b). Since the $IoU$ with $g_3$ is higher, we consider $p_1$ matching with $g_3$ and assign $g_3$'s relevance score to $p_1$. Then $g_3$ is removed from the ground truth due to being matched by a predicted moment. 
Assuming $p_2$ is very similar to $p_1$ (or a near duplicate),\footnote{This is a rare case, and it is unlikely to have two ground truth moments matching duplicate predictions as well. However, we would like to show that our proposed measure is able to handle such a rare case.} with $IoU(p_2, g_1)=0.35$ and $IoU(p_2, g_3)=0.4$, as depicted in Figure~\ref{fig:ndcgIoUAtk}(c). Since $g_3$ has been removed from the ground truth, $g_1$ becomes the sole match for $p_2$, resulting in a relevance score of 4 for $p_2$, and the removal of $g_1$ from the ground truth.
For $p_3$, $IoU(p_3, g_4)=0.5$, indicating a match with $g_4$, thus $p_3$ is assigned a relevance score of 2, and $g_4$ is removed.
The relevance scores of the predicted moments are 2, 4, and 2. To compute $NDCG$ of the ground truth ranking, we consider a perfect ranking of the top 3 ground truth moments (4, 2, and 2), regardless of whether they are matched by predicted moments or not, as illustrated by $K=3$ on the left side of Figure~\ref{fig:ndcgIoUAtk}(b).

 \section{Baseline Performance}
 \label{sec:bench_mark}
Illustrated in Figure~\ref{fig:taskComparison}, the closest task setting to RVMR is VCMR. In particular, if a VCMR model can compute a form of confidence for its retrieval result, then a ranking of the predicted moments can be easily achieved. Hence, we adapt  three VCMR models to RVMR and evaluate them on the TRV-Ranking :XML~\cite{lei2020tvr}, CONQUER~\cite{hou2021conquer}, and ReLoCLNet~\cite{zhang2021video}. The main change in the adaptations is the introduction of weight to the training loss based on query-moment similarity \ie $sim(q,m.c)$, recognizing that moments vary in relevance to a query. This weight aims to diminish the influence of less relevant moments on model training by adjusting the loss. The implementation details are in Appendix~\ref{subsec:baselinesImplem}.

\begin{table}[t]
    \centering
    \footnotesize
      \caption{Performance of the three baselines on the TVR-Ranking dataset. $N$ is the number of moments included in the pseudo training set for each query, by the query-caption similarity $sim(q, m.c)$. }
    \label{tab:expResultsN20}
    \begin{tabular}{ll|cccccc}
        \toprule
           & & \multicolumn{6}{c}{$NDCG@20$} \\ 
           \cmidrule{3-8}
           Model & $N$ & \multicolumn{2}{c}{$IoU \geq 0.3$} & \multicolumn{2}{c}{$IoU \geq 0.5$} & \multicolumn{2}{c}{$IoU \geq 0.7$} \\ 
           \cmidrule(lr){3-4}\cmidrule(lr){5-6}\cmidrule(lr){7-8}
           & & val & test & val & test & val & test  \\
        \midrule
        \multirow{3}{*}{XML~\cite{lei2020tvr}} 
            &  1 &  0.1010 & 0.0923 & 0.0737 & 0.0662 & 0.0258 & 0.0269  \\
            &  20 & 0.2331 & 0.2243 & 0.1700 & 0.1650 & 0.0627 & 0.0664  \\
            &  40 & 0.2114 & 0.2167 & 0.1530 & 0.1590 & 0.0583 & 0.0635  \\
        \midrule
        \multirow{3}{*}{CONQUER~\cite{hou2021conquer}} 
            & 1  & 0.0952 & 0.0835 & 0.0808 & 0.0687 & 0.0526 & 0.0484 \\
            & 20 & 0.2130 & 0.1995 & 0.1976 & 0.1867 & 0.1527 & 0.1368 \\
            & 40 & 0.2183 & 0.1968 & 0.2022 & 0.1851 & 0.1524 & 0.1365 \\
          \midrule
        \multirow{3}{*}{ReLoCLNet~\cite{zhang2021video}} 
            &  1 &  0.1504 & 0.1439 & 0.1303 & 0.1269 & 0.0866 & 0.0849  \\
            &  20 & 0.3815 & 0.3792 & 0.3462 & 0.3427 & 0.2381 & 0.2386  \\
            &  40 & 0.4418 & 0.4439 & 0.4060 & 0.4059 & 0.2787 & 0.2877  \\    
        \bottomrule
        \end{tabular}
   \end{table}

We conducted three sets of experiments with different $N=\{1, 20, 40\}$ values for a comprehensive evaluation, where $N$ is the number of moments included in the pseudo training set for each query, by the query-caption similarity $sim(q, m.c)$. We use evaluation metric $NDCG@K, IoU\geq \mu$, for $K=\{10, 20, 40\}$ and $\mu=\{ 0.3, 0.5, 0.7\}$. The parameter search is based on the best results on the validation set with $NDCG@20, IoU\geq 0.5$. Table~\ref{tab:expResultsN20} reports the experiment results with $NDCG@20$, and the full results are in Appendix~\ref{subsec:fullExpRes}. 

\paratitle{Performance on Test and Validation Sets}. We observe a consistent trend across all metrics, models, and pseudo-training sets: generally, the results on the test set exhibit slightly lower performance compared to the validation set, as expected. However, a few exceptions exist, but the differences in performance are marginal across all such cases.

\paratitle{The $K$ Values for $NDCG$.} When training with the top 1 pseudo training set, no clear pattern emerges across the three models as $K$ values change. With the top 20 and top 40 pseudo training sets, the $NDCG$ slightly increases as $K$ changes from 10 to 40 for XML and ReLoCLNet, while CONQUER shows the opposite trend.

\paratitle{The $\mu$ Values for $IoU$.} As expected, elevating the value of $\mu$ poses a greater challenge to localization, resulting in a decline in performance in general. The impact to models is a bit different as well. In particular,  XML  experiences a bigger drop compared to other models, suggesting its limitations in achieving precise localization. ReLoCNet shows relatively a smaller drop with higher  $\mu$ values.

\paratitle{The Choice of $N$ in Pseudo Training:} The $N=1$ training set yields the lowest scores across all three models, suggesting insufficient training instances. For XML, the best performance is achieved with $N=20$, while ReLoCLNet performs best with $N=40$. The results for CONQUER are comparable for $N=20$ and $N=40$. These findings indicate that the models have different capabilities in handling noise in the training data. 

\paratitle{Comparison of Baseline Models:} All three models (XML, CONQUER, and ReLoCNet) exhibit consistent performance trends across different training sets and metrics. Among the three, ReLoCNet emerges as the top performer, notably when sufficient training moments are provided \ie $N=40$. However, the three baselines show different performance ranking on the VCMR task, as reported in the original papers~\cite{lei2020tvr,hou2021conquer,zhang2021video}, where CONQUER demonstrates superior performance against ReLoCNet, and XML is slightly better than ReLoCNet as well. The discrepancy on RVMR implies that the abilities required by our RVMR task differ from those of the VCMR task.

Although VCMR models can be easily adapted, directly applying them to an RVMR application may not be appropriate.
Designing a new model tailored specifically to our RVMR task is necessary. These findings validate the significance and utility of our dataset for further research and development in the video retrieval field.

\section{Conclusion and Limitations}
\label{sec:conclude}
In this paper, we study the task of ranked moment retrieval to better reflect the practical setting of moment retrieval from video collection, by queries in natural language. To facilitate the research in this new task, we develop the \dataset based on the raw videos and moment annotations of the TVR dataset. Our data annotation process considers query rewriting to best simulate the queries from users who may not have watched all videos in the search collection. The main effort is the manual annotation of relevance levels for a large number of candidate moments for validation and test queries. We then develop the evaluation metric by considering measures used in both ranking tasks \ie $NDCG$, and in moment retrieval, \ie IoU. Through experiments, we show that models that perform well on VCMR may not necessarily outperform others on this new RVMR task, indicating the lack of ranking capability of existing models. 

Our work has three limitations. First, the queries used in the dataset might not perfectly mirror the real-world needs of users, potentially limiting their practical applicability. As the source videos are from six TV series, diversity is also a concern. Nonetheless, for the purpose of benchmarking and evaluating model capabilities, the dataset is adequate. Second, the annotation process employs a combination of query and caption as a proxy, identifying up to 40 relevant moments. This method may overlook some genuinely relevant moments, but make the annotation feasible for a reasonable coverage. Third, the pseudo training set is generated through a proxy $sim(q, m.c)$, which may lead to a gap compared to the real annotations in the validation and test sets. 

Our work provides relevance annotations on top of an existing open-source dataset. We do not anticipate any potential negative societal impacts.

{\small
\bibliographystyle{unsrt}
\bibliography{main.bib}
}

\newpage
\appendix

\section{Character Name Substitution}
\label{appendix:replacement_implementation}
\begin{figure}[hb]
    \centering
    \includegraphics[width=0.8\columnwidth]{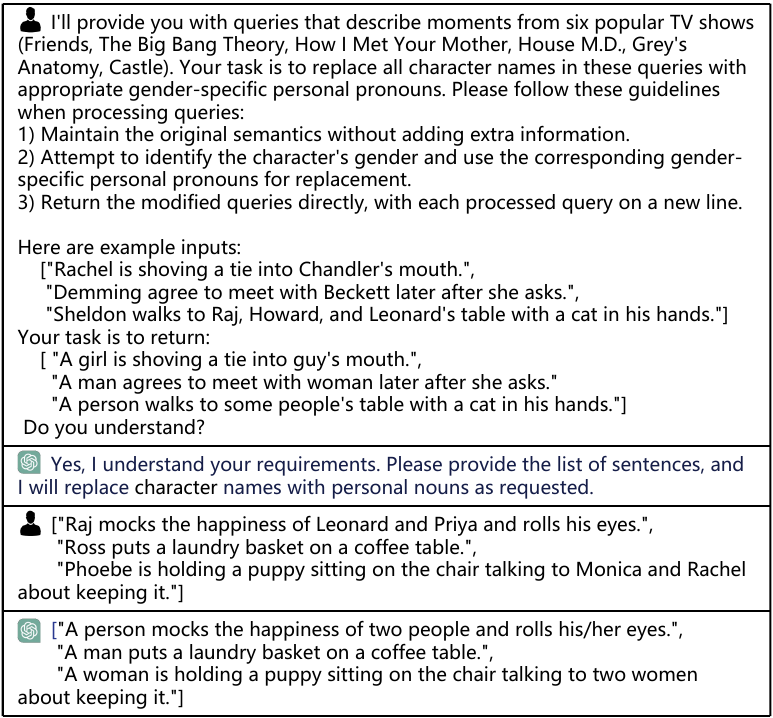}
    \caption{A prompt example and the subsequent conversation with ChatGPT for characters' names replacement.}
    \label{fig:chatgpt}
\end{figure}

We used ChatGPT 3.5 to substitute character name in moment descriptions from the original TVR dataset, through carefully designed prompts. Specifically, we request ChatGPT to replace all characters' names with appropriate personal pronouns in a given sentence.
We also instructed ChatGPT to estimate the character's gender according to its name, aiming at promoting the diversity of our queries.
An example conversation is shown in Figure~\ref{fig:chatgpt}.

Furthermore, we add a verification process to avoid potential errors in ChatGPT's responses and to ensure output quality.
The verification contains two steps. First, we assess the semantic alignment of the query before and after word substitution, using SimCSE~\cite{gao2021simcse} to calculate the sentence similarity. Second, we use Flair~\cite{akbik2019flair} to check if named entity recognition (NER) can be detected from the replaced version. Outputs deemed of inadequate quality are reprocessed by ChatGPT with a different random seed and subsequently reviewed by annotators.

The implementation details are as follows: The initial temperature parameter was set to 0 to minimize output variability. For reprocessing inadequate quality outputs, the temperature was adjusted to 0.2. During the sentence semantic assessment phase, approximately 0.04\% (42 queries) exhibited a similarity score below 0.4, failing the verification. In the NER check, around 0.01\% (15 queries) scored above 0.8, indicating unsatisfactory name replacement. After reprocessing and human review, all queries successfully met our expectations.

\section{Annotation Guideline, Annotator, and Annotation Analysis}
\label{appendix:secAnnotation}

\subsection{Annotation Guideline}
\label{appendix:annotationGuideline}

Based on annotated samples, we observed that annotators mainly consider three factors in judging the degree of relevances between candidate moments and queries: match of action(s), visual completeness, and temporal completeness. Action match indicates the alignment between query semantics and the action(s) demonstrated in the candidate moment. Visual completeness considers that the entire action is completed within the video frame. Temporal completeness evaluates the proportion of the complete action accounting for the moment. The relevance is low when too many unrelated segments appear at a moment, or when an action is incomplete or interrupted. Based on the observations, we design a guideline and use it for pre-annotation. We then refined the guideline based on the feedback received from the pre-annotation exercise. The formal annotation follows the guideline below to determine the degree of relevance from level 4 (perfect match) to level 0 (unrelated).

\begin{description}
    \item[\textbf{Level 4}] A moment perfectly matches the query if the actions expressed in the moment accurately align with the query's semantics. The action occurs at a prominent spot within the frames, and the timespan of the moment fully covers the action without redundancy.
    
    \item[\textbf{Level 3}] The moment could match the query well, except for a little point mismatched. The action generally aligns with the semantics of the query, but some details are missed.

    \item[\textbf{Level 2}] The moment  matches the search query, though there are a few noticeable mismatches. The action within the moment only conveys a part of the query information. There is a large gap in visual and temporal completeness. 

    \item[\textbf{Level 1}]  The moment presents relevance to the query on a few points. Though a few points of action are mentioned in the query, the main action is mismatched with the query.

    \item[\textbf{Level 0}] The moment is entirely unrelated to the query.
\end{description}

We provide examples to illustrate the annotation guidelines. For instance, if the query is "A person stands up and walks towards the board", and the moment shows one or two people standing up and walking towards the board, we consider it a perfect match and assign a relevance level of 4. Another moment might also reflect the query's semantics but include additional unrelated visual content; this moment will be assigned a relevance level of 3. If a moment only shows "a man walks towards a board" without the action "stand up," it will be assigned a relevance level of 2. An example of a level 1 moment might show "a man" and "a board" but no interaction between them. Note that, due to the nature of the video source as TV series, many moments will contain a "person". If none of their actions are mentioned in the query, such moments are considered unrelated. The "person" here is similar to stop words in web search.

\subsection{Annotation Setup and Annotators}
\label{appendix:annotator}

The original TVR dataset consists of 21,829 videos, each at a resolution of 480p and with a frame rate of 3 fps. These videos were segmented into 97,410 moments based on their timestamps and converted into GIF files, which were then stored on AWS S3 storage. For the annotation process, we selected Label Studio as our platform, which integrates well with our workflow. For each query, 20 candidate moments are initially identified through query-caption similarity. If more moments are needed, the next 20 candidate moments are identified and annotated. Consequently, the maximum number of moments that could be labeled per query is 40.

We received the number of 227,808 raw annotations from all annotators. After cleaning and merging, we obtained the number of 94,442 annotated moments for 3,281 queries with annotation consensus. 

Our annotation team consists of 23 diverse students from China, Singapore, and India, all of whom have undergraduate degrees or higher. The team maintains gender balance. The majority of our team members are native English speakers, and the rest possess strong English proficiency. This ensures that our annotators can accurately understand the queries in English.

All annotators underwent training including an explanation of the RVMR task, the annotation guidelines, and a tutorial of the annotation tool. This training ensures that all annotators share the same understanding of the task and follow the same annotation standards. After the training, all annotators took a qualification exam to align their standards. The qualification exam included 10 randomly sampled queries, with the top 10 candidate moments chosen based on query-caption similarity for each query. Annotators independently annotated these 100 query-moment pairs, and we collected their annotations. The median level of relevance from all annotations was set as the ground truth for each query-moment pair.

We followed the criteria in Section \ref{ssec:queryCollection} to measure consensus. Feedback was provided on all annotations that did not reach consensus, along with an explanation, and annotators were asked to relabel them. Initially, an average of 11.83\% of annotations failed to align with the ground truth. After feedback and relabeling, all annotations reached consensus. In the final exam results, the average Cohen's Kappa coefficient was 0.49. Given the subjective nature of determining the relevance between visual content and text, we consider this level of agreement to be reasonable. The workers are paid 14 SGD per hour under a part-time job contract. The annotation tool only records worker IDs and not their identities.

\subsection{Annotation Analysis: Relevance Level vs Moment-Caption Similarity}
\label{appendix:annotationAnalysis}

In the TVR-Ranking dataset, our goal is to find and label all the relevant moments for a given query. However, in the annotation process, locating all such moments is time-consuming and unnecessary. In the original TVR dataset, workers watched the source videos and wrote moment descriptions. Therefore, we use these moments as candidate moments for our annotation and consider the original moment descriptions to be a good proxy for visual relevance. To distinguish the moment descriptions before and after character name substitution, we refer to the substituted version as the \textit{moment caption}. We use $m.c$ to denote the moment caption and $m.v$ for the moment's visual content. All our annotations are based on moment captions. 

Given query-moment pairs, we study the relationship between the query-caption similarity $sim(q, m.c)$ and our annotated relevance between the query and the moment's visual content $rel(q, m.v)$. This study is based on annotations for 10 randomly sampled queries. Instead of annotating 40 candidate moments for each query, we have annotated 120 candidate moments for each of these 10 queries. Specifically, for each sampled query, we annotate the 60 candidate moments with the highest $sim(q, m.c)$ scores and then randomly sample another 60 moments from the remaining moments in the TVR dataset. 

\begin{figure}[th]
     \centering
     \begin{subfigure}[t]{0.33\linewidth}
         \centering
        \includegraphics[height=5cm]{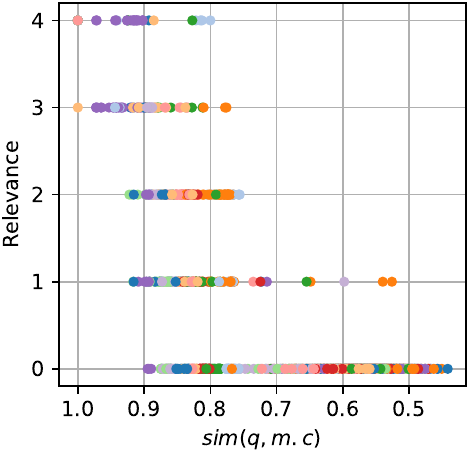}
        \caption{Relevance against $sim(q, m.c)$}
         \label{sfig:relBySim}
     \end{subfigure}
     \hspace{2cm}%
    \begin{subfigure}[t]{0.335\linewidth}
         \centering
         \includegraphics[height=5cm]{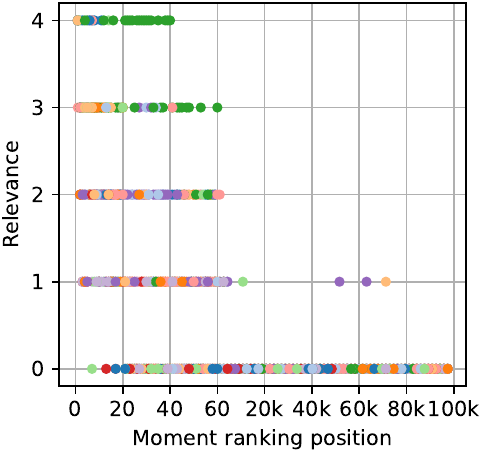}
        \caption{Against moment ranking position}
         \label{sfig:relByPosition}
     \end{subfigure}
        \caption{Relationship between relevance score (1 - 4) and $sim(q, m.c)$ with annotations for 10 sample queries. The candidate moments of the same query are in one color. In (b), moment ranking positions are by their similarity scores $sim(q, m.c)$ in descending order.} 
        \label{fig:relSimRelatoin}
\end{figure}

As shown in Figure~\ref{sfig:relBySim}, moments with higher $sim(q, m.c)$ scores are mostly annotated with high relevance scores. This suggests that moment captions, as human-generated descriptions of video segments, can be a good proxy for estimating a moment's relevance to a query. Therefore, our pseudo training set is expected to serve its purpose well, although it is not perfect. 

Figure~\ref{sfig:relByPosition} illustrates the relevance scores of candidate moments against their ranking positions by $sim(q, m.c)$. Recall that for each query, we included the top 60 candidates by $sim(q, m.c)$ and another randomly sampled 60 candidates, which may have very low ranking positions by $sim(q, m.c)$ (1 being the highest ranking position). As shown in the figure, moments with high relevance scores (e.g., 4 and 3) mostly have ranking positions higher than 40, and very few relevant moments are ranked beyond the position of 60. This figure also demonstrates the correlation between the relevance score $rel(q, m.v)$ and $sim(q, m.c)$. More importantly, it strongly supports our choice of annotating the top 40 candidate moments by $sim(q, m.c)$, which strikes a good balance between capturing relevant moments for a query and managing the annotation load.

\subsection{Annotation Analysis: Annotation Consensus and Distribution}

\begin{figure}
     \centering
     \begin{subfigure}[b]{0.35\linewidth}
         \centering
         \includegraphics[height=5cm]{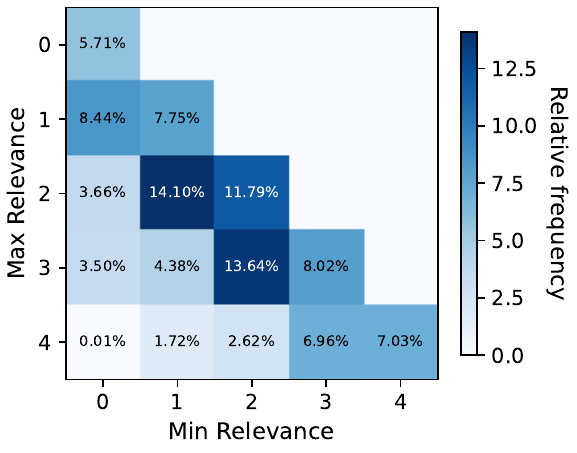}
         \caption{Raw annotation scores}
         \label{sfig:rawAnnotations}
     \end{subfigure}
    \hspace{2cm}%
     \begin{subfigure}[b]{0.30\linewidth}
         \centering
         \includegraphics[height=5cm]{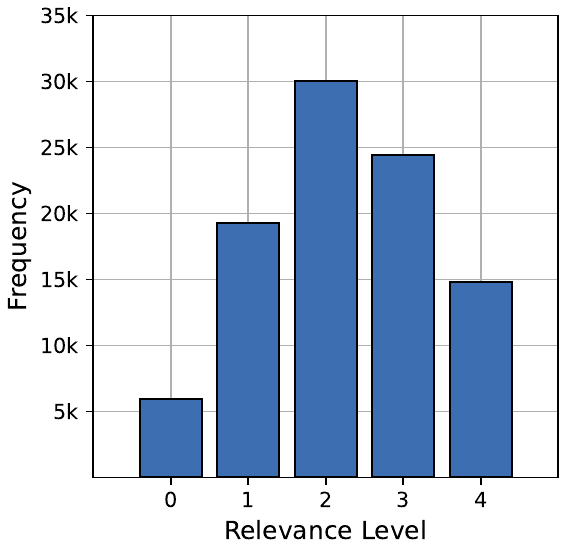}
         \caption{Frequency of final scores}
         \label{sfig:finalScores}
     \end{subfigure}
     \hfill
        \caption{(a) Distribution of the relevant scores in all raw annotations by two or four annotators. (b) Distribution of the final scores after discarding annotations in disagreement.}
        \label{fig: relevance similarity}
   
\end{figure}

To elaborate on the level of consensus among the annotators, we show the distribution of the raw scores they assigned. Recall that each query-moment pair is annotated by either two annotators (if they reach a consensus) or four annotators (if a consensus is not reached initially). Thus, for each query-moment pair, we have either 2 or 4 raw scores. These scores range from a minimum of 0 (non-relevant) to a maximum of 4 (perfect match). We group query-moment pairs according to their minimum and maximum scores among the raw scores assigned by the annotators. Figure~\ref{sfig:rawAnnotations} shows the distribution of all query-moment pairs. For instance, the cell at the bottom left corner represents the percentage of query-moment pairs with a minimum score of 0 and a maximum score of 4, which is extremely rare at 0.01\%, or one in ten thousand, likely due to human errors during the annotation process. The majority of query-moment pairs have either identical scores assigned by all annotators or a small difference of 1 between the minimum and maximum scores, indicating a consensus.

In Figure~\ref{sfig:finalScores}, we present the distribution of final relevance scores for all query-moment pairs. 
The final score is calculated as the mean score from annotators, rounded up to the nearest whole number, or as the trimmed mean if four annotators were involved. The distribution roughly follows a normal distribution, with fewer pairs exhibiting relevance scores of 0 or 4, and the majority having a relevance score of 2. Note that for each query, we annotate at most 40 candidate moments. A relevance score of 0 indicates confirmed non-relevance, which differs from moments that are not annotated for the query, although the latter are also likely to be irrelevant. 

\subsection{Annotation Analysis: Case Study of Example Queries}
\label{ssec:dataQulity}

\begin{figure*}
    \centering
    \includegraphics[width=0.9\textwidth]{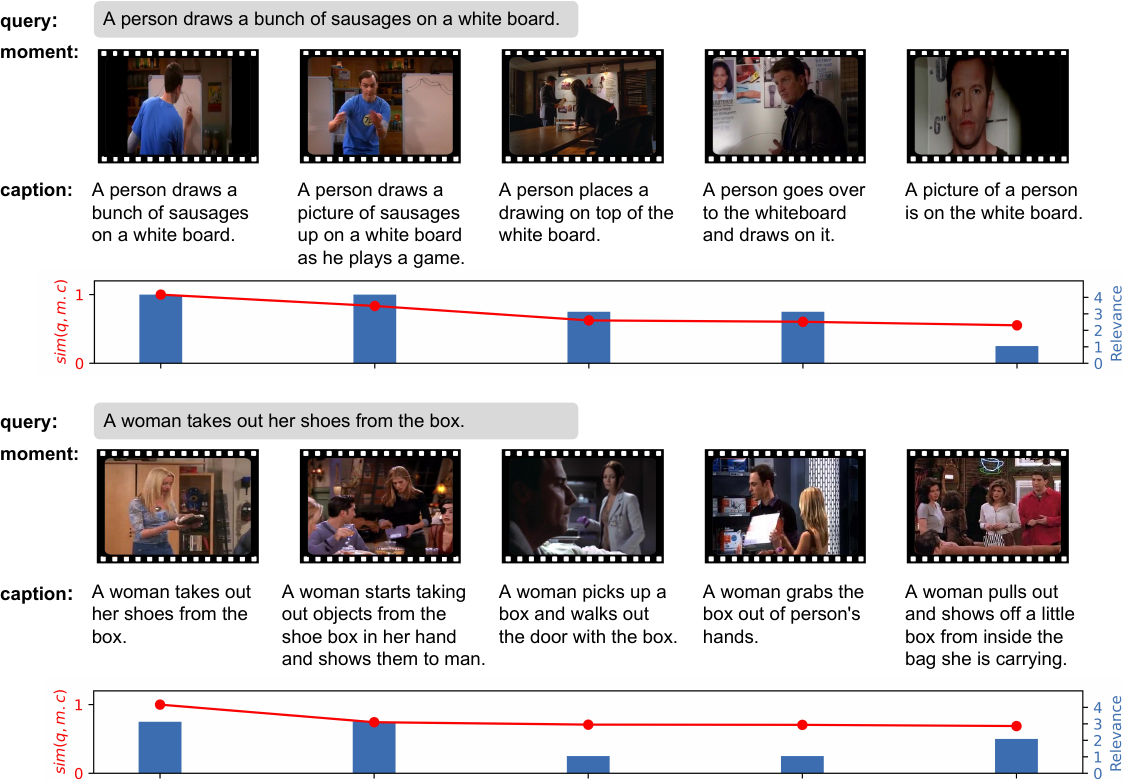}
    \caption{Two queries with candidate moments ranked by $sim(q,m.c)$ in the line chart, and final relevance scores in the bar chart.}
    \label{fig:query_multiple_moments}
\end{figure*}

Figure~\ref{fig:query_multiple_moments} provides two example queries, each with 5 candidate moments ranked by $sim(q, m.c)$ in descending order. The relevance scores assigned by annotators (in bar chart) show a reasonable correlation with $sim(q, m.c)$ (in line chart), where larger $sim(q, m.c)$ also suggests high relevance scores. This is expected based on the earlier analysis in Figure~\ref{sfig:relBySim}. Yet, the moments with lower $sim(q, m.c)$'s can be annotated with higher relevance scores, and $sim(q, m.c)=1.0$ does not guarantee a perfect match. The second example is a good illustration of this point. 
One possible reason for the discrepancy is the annotation methodology. The original TVR dataset required annotators to watch the full video before selecting and describing specific moments, providing them with full context in the video. In our annotation, only moments are presented to annotators, without the full video. Our annotators make judgments solely based on the provided moment. For query ``A woman takes out her shoes from the box'', the moment with $sim(q, m.c)=1$ is not considered a perfect match because the moment does not show ``the box'', though ``the box'' might exist in the source video somewhere before this moment. Among the test query set, there are 2,635 queries where annotators reach consensus for the query moment pair with $sim(q, m.c)=1.0$. Among them, 312 (11.76\%) moments are not assigned with the perfect relevance score.

We clarify two points here. First, each query is a moment caption originated from the TVR dataset, hence there exists at least one candidate moment whose caption is identical to the query, \ie $sim(q, m.c)=1.0$. Second, the moment captions in Figure~\ref{fig:query_multiple_moments} are provided for reference purposes here. The captions of candidate moments are not shown to the annotators during the annotation process. Annotators judge the level of relevance purely based on moment's visual content to the query. Further,  moment captions in the dataset shall only be used as queries, and not as additional information available in source videos. In a RMVR task, the videos are not segmented into moments and such high-quality captions do not exist as well.

\section{Experiments and Case Study on Evaluation}
\label{appendix:ExpCaseStudy}

\begin{table}[th]
    \centering
    \footnotesize
      \caption{Performance of the three baselines on the TVR-Ranking dataset. $N$ is the number of moments included in the pseudo training set for each query, by the query-caption similarity $sim(q, m.c)$. }
    \label{tab:expResults}
    
    \begin{subtable}{1.0\linewidth}
        \centering
    \begin{tabular}{ll|cccccc}
        \toprule
           & & \multicolumn{6}{c}{$NDCG@10$} \\ 
           \cmidrule{3-8}
           Model & $N$ & \multicolumn{2}{c}{$IoU \geq 0.3$} & \multicolumn{2}{c}{$IoU \geq 0.5$} & \multicolumn{2}{c}{$IoU \geq 0.7$} \\ 
           \cmidrule(lr){3-4}\cmidrule(lr){5-6}\cmidrule(lr){7-8}
           & & val & test & val & test & val & test  \\
        \midrule
        \multirow{3}{*}{XML~\cite{lei2020tvr}} 
            &  1  &  0.1016 & 0.0917 & 0.0747 & 0.0660 & 0.0244 & 0.0268 \\
            &  20 &  0.2226 & 0.2135 & 0.1623 & 0.1567 & 0.0580 & 0.0627 \\
            &  40 &  0.2002 & 0.2044 & 0.1461 & 0.1502 & 0.0541 & 0.0589 \\
        \midrule
        \multirow{3}{*}{CONQUER~\cite{hou2021conquer}} 
            & 1  & 0.0999 & 0.0859 & 0.0844 & 0.0709 & 0.0530 & 0.0512 \\
            & 20 & 0.2406 & 0.2249 & 0.2222 & 0.2104 & 0.1672 & 0.1517 \\
            & 40 & 0.2450 & 0.2219 & 0.2262 & 0.2085 & 0.1670 & 0.1515 \\
          \midrule
        \multirow{3}{*}{ReLoCLNet~\cite{zhang2021video}} 
            &  1  &  0.1575 & 0.1525 & 0.1358 & 0.1349 & 0.0908 & 0.0916 \\
            &  20 &  0.3751 & 0.3751 & 0.3407 & 0.3397 & 0.2316 & 0.2338 \\
            &  40 &  0.4339 & 0.4353 & 0.3984 & 0.3986 & 0.2693 & 0.2807 \\
        \bottomrule
        \end{tabular}
    \end{subtable}%
    \newline
    \vspace{1ex}
    \begin{subtable}{1.0\linewidth}
        \centering
    \begin{tabular}{ll|cccccc}
        \toprule
           & & \multicolumn{6}{c}{$NDCG@20$} \\ 
           \cmidrule{3-8}
           Model & $N$ & \multicolumn{2}{c}{$IoU \geq 0.3$} & \multicolumn{2}{c}{$IoU \geq 0.5$} & \multicolumn{2}{c}{$IoU \geq 0.7$} \\ 
           \cmidrule(lr){3-4}\cmidrule(lr){5-6}\cmidrule(lr){7-8}
           & & val & test & val & test & val & test  \\
        \midrule
        \multirow{3}{*}{XML} 
            &  1  &  0.1010 & 0.0923 & 0.0737 & 0.0662 & 0.0258 & 0.0269 \\
            &  20 &  0.2331 & 0.2243 & 0.1700 & 0.1650 & 0.0627 & 0.0664 \\
            &  40 &  0.2114 & 0.2167 & 0.1530 & 0.1590 & 0.0583 & 0.0635 \\
        \midrule
        \multirow{3}{*}{CONQUER} 
            & 1  & 0.0952 & 0.0835 & 0.0808 & 0.0687 & 0.0526 & 0.0484 \\
            & 20 & 0.2130 & 0.1995 & 0.1976 & 0.1867 & 0.1527 & 0.1368 \\
            & 40 & 0.2183 & 0.1968 & 0.2022 & 0.1851 & 0.1524 & 0.1365 \\
          \midrule
        \multirow{3}{*}{ReLoCLNet\hspace{6mm}} 
            &  1  &  0.1504 & 0.1439 & 0.1303 & 0.1269 & 0.0866 & 0.0849 \\
            &  20 &  0.3815 & 0.3792 & 0.3462 & 0.3427 & 0.2381 & 0.2386 \\
            &  40 &  0.4418 & 0.4439 & 0.4060 & 0.4059 & 0.2787 & 0.2877 \\
        \bottomrule
        \end{tabular}
    \end{subtable}
    \newline
    \vspace{1ex}
    \begin{subtable}{1.0\linewidth}
        \centering
    \begin{tabular}{ll|cccccc}
        \toprule
            & & \multicolumn{6}{c}{$NDCG@40$} \\ 
           \cmidrule{3-8}
           Model & $N$ & \multicolumn{2}{c}{$IoU \geq 0.3$} & \multicolumn{2}{c}{$IoU \geq 0.5$} & \multicolumn{2}{c}{$IoU \geq 0.7$} \\ 
           \cmidrule(lr){3-4}\cmidrule(lr){5-6}\cmidrule(lr){7-8}
           & & val & test & val & test & val & test  \\
        \midrule
        \multirow{3}{*}{XML} 
            &  1  &  0.1077 & 0.1016 & 0.0775 & 0.0727 & 0.0273 & 0.0294 \\
            &  20 &  0.2580 & 0.2512 & 0.1874 & 0.1853 & 0.0705 & 0.0753 \\
            &  40 &  0.2408 & 0.2432 & 0.1740 & 0.1791 & 0.0666 & 0.0720 \\
        \midrule
        \multirow{3}{*}{CONQUER} 
            & 1  & 0.0974 & 0.0866 & 0.0832 & 0.0718 & 0.0557 & 0.0510 \\
            & 20 & 0.2029 & 0.1906 & 0.1891 & 0.1788 & 0.1476 & 0.1326 \\
            & 40 & 0.2080 & 0.1885 & 0.1934 & 0.1775 & 0.1473 & 0.1323 \\
          \midrule
        \multirow{3}{*}{ReLoCLNet\hspace{6mm}} 
            &  1  &  0.1533 & 0.1489 & 0.1321 & 0.1304 & 0.0878 & 0.0869 \\
            &  20 &  0.4039 & 0.4031 & 0.3656 & 0.3648 & 0.2542 & 0.2567 \\
            &  40 &  0.4725 & 0.4735 & 0.4337 & 0.4337 & 0.3015 & 0.3079 \\
            \bottomrule
        \end{tabular}
    \end{subtable}
\end{table}

In our experiments, we adapt three VCMR models to RVMR and evaluate them on the TRV-Ranking: XML~\cite{lei2020tvr}, CONQUER~\cite{hou2021conquer}, and ReLoCLNet~\cite{zhang2021video}. The main change in the adaptations is the introduction of weight to the training loss based on query-moment similarity, \ie $sim(q,m.c)$, recognizing that moments vary in relevance to a query. There is no such loss in the VCMR task setting because there is exactly one ground truth in VCMR. This weight aims to diminish the influence of less relevant moments on model training by adjusting the loss. In terms of features, we follow the original implementation and use both subtitle and video features extracted from the source video for all three models. The query features in our implementation are extracted using BERT~\cite{devlin2018bert}.

Note that, because VCMR was directly extended from VMR, there is a train/test split in the source videos. That is, there is a set of queries and source videos for training, and another set for testing in VCMR. However, the RVMR task is similar to web search, except the result documents are video moments. In the RVMR task setting, both training and test queries search for moments from a common large pool of source videos. As long as there are no duplicates between training and test queries, there is no data leakage issue.

\subsection{Baselines and Implementation Details}
\label{subsec:baselinesImplem}

\paratitle{Cross-modal Moment Localization (XML)}~\cite{lei2020tvr}.
The XML model was proposed alongside the TVR dataset. The motivation behind XML is to consider both video and subtitle information when retrieving moments, as some queries in TVR are based on subtitles. The model integrates video and subtitle features as context information and conducts retrieval on this context to achieve more accurate recall.
During the retrieval stage, the authors use matrix multiplication to compute the confidence score, enhancing retrieval efficiency. Additionally, they designed another branch specifically to predict the start and end times of moments. The XML model is licensed under the MIT license.

The video features in XML combine visual features from ResNet~\cite{he2016deep} and temporal features from I3D~\cite{carreira2017quo}. The subtitle features are extracted via RoBERTa~\cite{liu2019roberta}.
For the loss, we consider two factors. First, we multiply the similarity \( \text{sim}(q,m.c) \) (ranging from 0.0 to 1.0) in the original loss as a decay coefficient. We expect moments that are more likely to be relevant to have a greater impact on the model during training. Additionally, XML employs a video-level contrastive loss, pulling positive videos closer and pushing negative videos farther away. However, in our task setting, there could be multiple positive videos within a mini-batch for one query, because in the pseudo training set all top $N$ videos can be regarded as positive samples.
To address the issue, we introduce a \textit{positive pair mask}~\cite{khosla2020supervised}. This mask assigns a value of one to positions where the query and video are related in the pseudo training set, and zeros out other positions. When computing the loss, we sum the scores masked by the positive pair mask instead of using only the diagonal elements as the nominator.

\paratitle{Retrieval and Localization Network with Contrastive Learning (ReLoCLNet)}~\cite{zhang2021video}. 
The structure of ReLoCLNet is similar to XML. Both models integrate video and subtitle features as context, separately encode query and context embeddings, and use video-level contrastive loss. The key difference is that ReLoCLNet adds a frame-level contrastive loss to highlight the ground truth region and improve recall accuracy. The model is under the MIT license. 

In our implementation, ReLoCLNet uses the same video, subtitle, and query features as XML. The loss modifications are also similar to those in XML. We do not need to modify the frame-level contrastive loss because it does not involve the issue of multiple positive pairs.

\paratitle{CONtextual QUery-awarE Ranking (CONQUER)}~\cite{hou2021conquer}.
CONQUER uses a different framework compared to the other two models. It first retrieves candidate videos from the corpus and then conducts moment localization within those candidate videos.

CONQUER uses video features that concatenate the SlowFast~\cite{feichtenhofer2019slowfast} and ResNet models, while subtitle features are extracted using RoBERTa~\cite{liu2019roberta}. Additionally, CONQUER leverages video retrieval results from HERO~\cite{li2020hero}. We continue to use these results when training on our TVR-Ranking dataset. Note that, because the HERO results are obtained from the TVR dataset, there could be a data leak issue in our task setting. However, this issue is negligible for two reasons: (i) the queries used in our setting is imprecise query with query re-written, and (ii) a query has multiple ground truth moments in our task setting, which was not annotated in the original TVR dataset.

For all three models, we set the learning rate to 0.0001 with a warmup phase. For the $N=1$ pseudo training set, we trained the models for 4000 epochs, evaluating the validation set every 20 epochs. For the $N=20$ and $N=40$ training sets, we trained for 200 and 100 epochs, respectively, and evaluated the validation set 1 and 2 times per epoch, respectively. We used early stopping, set to trigger if performance did not improve after 10 evaluations. We adhered to the original model parameters and did not use non-maximum suppression for post-processing the outputs. All models were adapted to the same PyTorch version (2.2.1) with CUDA 12.1. The experiments were conducted on a single NVIDIA V100 32GB GPU.

\subsection{Experiment Results}
\label{subsec:fullExpRes}

Table~\ref{tab:expResults} reports the full results, conducted with different $N=\{1, 20, 40\}$ values, and evaluated by $NDCG@K, IoU\geq \mu$, for $K=\{10, 20, 40\}$ and $\mu=\{ 0.3, 0.5, 0.7\}$. The parameter search is based on the best results on the validation set with $NDCG@20, IoU\geq 0.5$. We have also conducted three runs of the ReLoCLNet model using different seeds with the top 40 pseudo training set. As reported in Table~\ref{tab:expDiffSeed}, the results are highly consistent with very small standard deviations

\begin{table}
    \centering
    \scriptsize 
    \setlength{\tabcolsep}{3pt} 
    \caption{Average $NDCG@K, IoU=\mu$ scores and their standard deviations of the ReLoCLNet model. Results are obtained from three runs of the model using different seeds, with the top 40 pseudo training set.}
    \label{tab:expDiffSeed}
    \begin{subtable}{1.0\linewidth}
        \centering
    \begin{tabular}{c|*{6}{>{\centering\arraybackslash}p{1.9cm}}}
        \toprule
            $NDCG$ & \multicolumn{2}{c}{$IoU \geq 0.3$} & \multicolumn{2}{c}{$IoU \geq 0.5$} & \multicolumn{2}{c}{$IoU \geq 0.7$} \\ 
           \cmidrule(lr){2-3}\cmidrule(lr){4-5}\cmidrule(lr){6-7}
            @ & val & test & val & test & val & test  \\
            \midrule
            \cmidrule{2-7}
               10 &  $0.4287 \pm 0.0048$ & $0.4347 \pm 0.0008$ & $0.3943 \pm 0.0038$ & $0.3982 \pm 0.0009$ & $0.2696 \pm 0.0029$ & $0.2809 \pm 0.0006$  \\ 
               20 &  $0.4359 \pm 0.0053$ & $0.4425 \pm 0.0010$ & $0.4010 \pm 0.0045$ & $0.4050 \pm 0.0008$ & $0.2761 \pm 0.0033$ & $0.2871 \pm 0.0010$ \\
               40 &  $0.4668 \pm 0.0050$ & $0.4724 \pm 0.0008$ & $0.4290 \pm 0.0041$ & $0.4326 \pm 0.0008$ & $0.2983 \pm 0.0041$ & $0.3077 \pm 0.0007$ \\
        \bottomrule
        \end{tabular}
    \end{subtable}%
\end{table}

To explore the difference between the VCMR task and our RVMR task, we compare these models' performance on both tasks.
The results on VCMR are summarized in Table~\ref{tab:expResultsVCMR}, taken from the original papers ~\cite{lei2020tvr,hou2021conquer,zhang2021video}. 
For the VCMR task, CONQUER demonstrates the best performance on the VCMR task, surpassing ReLoCLNet by 3.61\% with $IoU \geq 0.7$ and $R=1$, and by 8.40\% with $IoU \geq 0.7$ and $R=10$. 
However, CONQUER falls behind ReLoCLNet on the RVMR task for test set, with $NDCG@20$ lower by 0.2471, 0.2208, 0.1512 for $IoU \geq {0.3, 0.5, 0.7}$ respectively when using top 40 pseudo training set.
XML performs well on the VCMR task as well, achieving 44.44\% with $IoU \geq 0.5$ and $R=100$, slightly outperforming ReLoCLNet's 44.10\%.
However, it showed much poorer performance on the RVMR task. The discrepancy implies that the abilities required for the RVMR task differ from those for the VCMR task. Although VCMR models can be easily adapted, directly applying them to an RVMR application may not be appropriate. Thus, designing a new model tailored specifically to our RVMR task is necessary.

\begin{table}
    \centering
    \footnotesize
    \caption{Performance of XML, CONQUER, and ReLoCLNet on the TVR dataset for VCMR task. All values are taken from their original papers.}
    \label{tab:expResultsVCMR}
    \begin{tabular}{l|cccccc}
        \toprule
            & \multicolumn{3}{c}{$IoU \geq 0.5$} & \multicolumn{3}{c}{$IoU \geq 0.7$} \\ 
           \cmidrule(lr){2-4} \cmidrule(lr){5-7}
           Model                    & $R=1$ & $R=10$ & $R=100$  & $R=1$ & $R=10$ & $R=100$ \\
            \midrule
            XML~\cite{lei2020tvr}              & 7.25 & 21.65 & 44.44 & 3.25 & 12.49 & 29.51 \\
            CONQUER~\cite{hou2021conquer}      & -    & -     & -     & 7.76 & 22.49 & 35.17 \\
            ReLoCLNet~\cite{zhang2021video}    & 8.03 & 21.37 & 44.10 & 4.15 & 14.06 & 32.42 \\
            \bottomrule
        \end{tabular}
\end{table}

\subsection{Case study of Predictions for Sample Queries}

\begin{figure*}[t]
     \centering
     \begin{subfigure}[t]{0.31\linewidth}
         \centering
         \includegraphics[height=6.8cm]{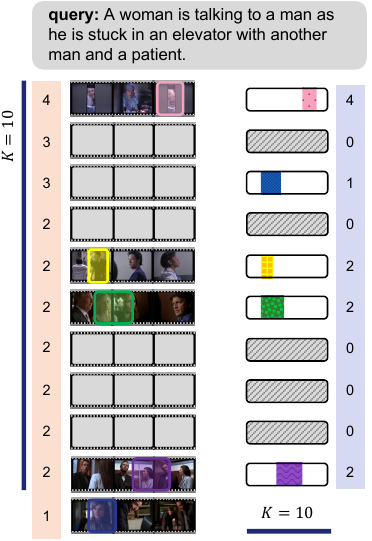}
         \caption{Result = 0.6167.}
         \label{fig:pred_case1}
     \end{subfigure}
     \hspace{0.2cm}
    \begin{subfigure}[t] {0.31\linewidth}
         \centering
         \includegraphics[height=6.8cm]{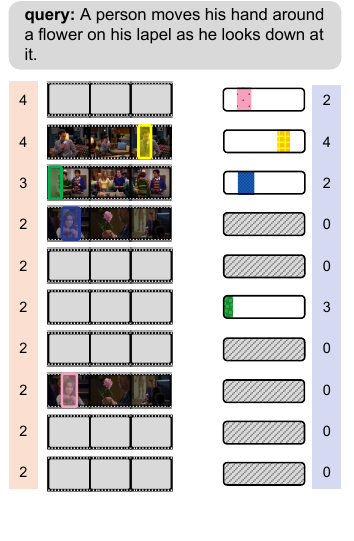}
         \caption{Result= 0.4675.}
         \label{fig:pred_case2}
     \end{subfigure}
     \hspace{-0.1cm}
    \begin{subfigure}[t]{0.31\linewidth}
         \centering
         \includegraphics[height=6.8cm]{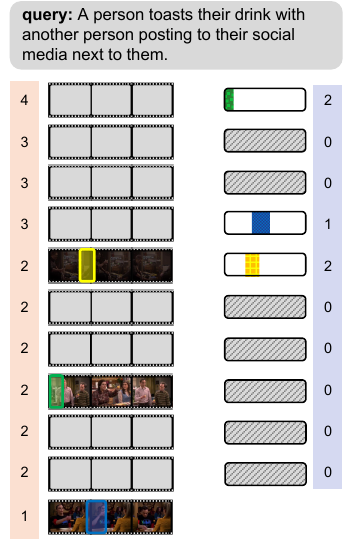}
         \caption{Result=0.1523.}
         \label{fig:pred_case3}
     \end{subfigure}
        \caption{
        Three example queries, along with their ground truth moments (on the left) and the predicted moments (on the right) with relevance levels indicated on their sides. The predicted results are measured by $NDCG@10, IoU \geq 0.3$. To clearly show the mapping of ground truth and the predicted moments, we mask the ground truth that are missed by the model. The  masked videos along the prediction column means incorrect predictions \ie not matching any ground truth.
        }
    \label{fig:prediction_case_study}    
\end{figure*}

To provide a more intuitive understanding of the RVMR task and evaluation, we present three samples with different scores in Figure~\ref{fig:prediction_case_study}. These queries are from the test set and the results are by ReLoCLNet when using top 40 pseudo training set. The predicted results are measured by $NDCG@10, IoU \geq 0.3$. Ground truth moments and the predicted moments (with relevance levels indicated on the side) are presented in the left and right columns respectively, for each query.  To clearly show the mapping of ground truth and the predicted moments, we mask the ground truth moments that are missed by the model. The masked videos along the prediction column means incorrect predictions \ie not matching any ground truth.

The first case shows a high $NDCG$ score as the model successfully finds the most relevant moment for the query, along with some other relevant moments.
The second case shows a similar sets of predicted relevant moments. However, this query has a few highly relevant ground truth moments, leading to a $NDCG$ score close to 0.5. 
The third case is low-score example where the model retrieves three relatively low relevant moments, missing the highly relevant ones.

\end{document}